\documentclass{article}

\usepackage{amsmath,graphicx,subfig}
\usepackage{amssymb,amsfonts}
\usepackage{overpic}
\usepackage[cal=boondox]{mathalpha}
\usepackage{bm}
\usepackage{enumitem}
\graphicspath{ {./images/} }
\usepackage{multirow}
\usepackage{array}
\usepackage{mdwmath}
\usepackage{mdwtab}
\usepackage{booktabs}
\usepackage[nodisplayskipstretch]{setspace} 
\usepackage[ruled,vlined,lined,commentsnumbered]{algorithm2e}
\usepackage[compact]{titlesec}
\newcommand\blfootnote[1]{%
  \begingroup
  \renewcommand\thefootnote{}\footnote{#1}%
  \addtocounter{footnote}{-1}%
  \endgroup
}

\usepackage{arxiv}

\usepackage[utf8]{inputenc} 
\usepackage[T1]{fontenc}    
\usepackage{hyperref}       
\usepackage{url}            
\usepackage{booktabs}       
\usepackage{amsfonts}       
\usepackage{nicefrac}       
\usepackage{microtype}      
\usepackage{lipsum}		
\usepackage{graphicx}
\usepackage{doi}

\usepackage{tikz}
\usepackage{textcomp}
\newcommand\copyrighttext{%
\footnotesize \textcopyright 2022 IEEE. Personal use of this material is permitted.
Permission from IEEE must be obtained for all other uses, in any current or future
media, including reprinting/republishing this material for advertising or promotional
purposes, creating new collective works, for resale or redistribution to servers or
lists, or reuse of any copyrighted component of this work in other works.

\doi{10.1109/ICIP46576.2022.9897190}}
\newcommand\copyrightnotice{%
\begin{tikzpicture}[remember picture,overlay]
\node[anchor=south,yshift=20pt] at (current page.south) {\fbox{\parbox{\dimexpr\textwidth-\fboxsep-\fboxrule\relax}{\copyrighttext}}};
\end{tikzpicture}%
}

\title{Efficient CNN-based Super Resolution Algorithms for mmWave Mobile Radar Imaging}

\author{Christos Vasileiou \\
	Department of Electrical and Computer Engineering\\
	The University of Texas at Dallas\\
	Richardson, TX 75080 \\
	\texttt{christos.vasileiou@utdallas.edu} \\
	\And
        \href{https://orcid.org/0000-0002-3388-4805}{\includegraphics[scale=0.06]{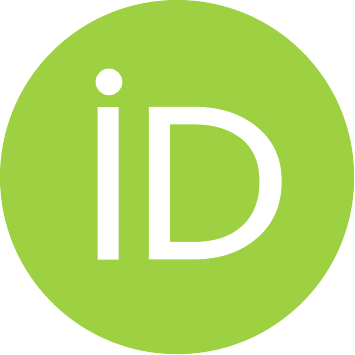}\hspace{1mm}Josiah W. Smith} \\
	Department of Electrical and Computer Engineering\\
	The University of Texas at Dallas\\
	Richardson, TX 75080 \\
	\texttt{josiah.smith@utdallas.edu} \\
	\And
	\href{https://orcid.org/0000-0002-0427-3773}{\includegraphics[scale=0.06]{orcid.pdf}\hspace{1mm}Shiva Thiagarjan} \\
	Department of Electrical and Computer Engineering\\
	The University of Texas at Dallas\\
	Richardson, TX 75080 \\
        \texttt{shivashankar.thiagarajan@utdallas.edu}
	\And
	  Matthew Nigh \\
	Department of Electrical and Computer Engineering\\
	The University of Texas at Dallas\\
	Richardson, TX 75080 \\
	\texttt{matthew.nigh@utdallas.edu} \\
	\And
	\href{https://orcid.org/0000-0002-4322-0068}{\includegraphics[scale=0.06]{orcid.pdf}\hspace{1mm}Yiorgos Markis} \\
	Department of Electrical and Computer Engineering\\
	The University of Texas at Dallas\\
	Richardson, TX 75080 \\
	\texttt{yiorgos.makris@utdallas.edu} \\
	\And
	\href{https://orcid.org/0000-0001-7229-1765}{\includegraphics[scale=0.06]{orcid.pdf}\hspace{1mm}Murat Torlak}\thanks{The work of Murat Torlak (while serving at NSF) was supported by NSF.} \\
	Department of Electrical and Computer Engineering\\
	The University of Texas at Dallas\\
	Richardson, TX 75080 \\
	\texttt{torlak@utdallas.edu} \\
}

\date{}

\renewcommand{\headeright}{\textit{IEEE Proc. ICIP} (Accepted)}
\renewcommand{\undertitle}{\textit{IEEE Proc. ICIP} (Accepted)}
\renewcommand{\shorttitle}{Efficient CNN-based Super-Resolution}

\hypersetup{
pdftitle={Efficient CNN-based Super Resolution Algorithms for mmWave Mobile Radar Imaging},
pdfsubject={cs.CV, cs.AI, eess.SP},
pdfauthor={Christos Vasileiou, Josiah W.~Smith, Shiva Thiagarajan, Matthew Nigh, Yiorgos Makris, Murat Torlak},
pdfkeywords={Fast-Freehand Imaging, Super-Resolution Algorithms, Adversarial Networks, mmWave(5G) Mobile Imaging, Depth-wise Convolution},
}

\begin{document}
\maketitle
\copyrightnotice

\begin{abstract}
In this paper, we introduce an innovative super resolution approach to emerging modes of near-field synthetic aperture radar (SAR) imaging. 
Recent research extends convolutional neural network (CNN) architectures from the optical to the electromagnetic domain to achieve super resolution on images generated from radar signaling. 
Specifically, near-field synthetic aperture radar (SAR) imaging, a method for generating high-resolution images by scanning a radar across space to create a synthetic aperture, is of interest due to its high-fidelity spatial sensing capability, low cost devices, and large application space.
Since SAR imaging requires large aperture sizes to achieve high resolution, super-resolution algorithms are valuable for many applications.
Freehand smartphone SAR, an emerging sensing modality, requires irregular SAR apertures in the near-field and computation on mobile devices.
Achieving efficient high-resolution SAR images from irregularly sampled data collected by freehand motion of a smartphone is a challenging task. 
In this paper, we propose a novel CNN architecture to achieve SAR image super-resolution for mobile applications by employing state-of-the-art SAR processing and deep learning techniques.  
The proposed algorithm is verified via simulation and an empirical study. 
Our algorithm demonstrates high-efficiency and high-resolution radar imaging for near-field scenarios with irregular scanning geometries. 
\end{abstract}

\blfootnote{This work was supported in part by Texas Instruments through the Foundational Technology Research Centre and the Texas Analog Center of Excellence. The authors acknowledge the Texas Advanced Computing Center (TACC) at The University of Texas at Austin for providing HPC resources that have contributed to the research results reported within this paper. URL: http://www.tacc.utexas.edu}


\section{Introduction}
\label{sec:intro}

With the advancements in affordable handheld wireless devices, the implementation of mmWave radars on mobile platforms have become feasible over the last few years. Being safe for human use and power conscious, mmWave sensing has emerged in a wide variety of applications like gesture sensing \cite{smith2021sterile} and medical imaging \cite{fedeli2020microwave}. 
Today, many of these applications require dedicated scanning platforms for collecting high quality radar images. 
With the convenience and efficiency provided by the mobile platforms, these applications can be expanded to simplify the data collection process while it comes with the additional trade-offs on image quality and power consumption.

Recent research has enabled data collection using freehand near-field synthetic aperture radar (SAR) using positioning sensors commonly found in smartphones \cite{alvarez2021towards}. 
Freehand mmWave imaging is a technique to recover mmWave images by scanning a radar throughout space by hand. 
Compared to traditional near-field SAR techniques \cite{yanik2019sparse,smith2020nearfieldisar}, freehand imaging arrays do not conform to conventional geometries due to the arbitrary motion of the user's hand. 
Previous work \cite{alvarez2021towards} on freehand imaging employs the computationally prohibitive back projection algorithm (BPA). 
The range migration algorithm (RMA) \cite{yanik2019sparse} is an efficient Fourier-based implementation of the BPA, but it requires strictly co-planar samples and is not suitable for freehand imaging.
Recent work \cite{smith2022efficient} proposes an efficient RMA-based algorithm that projects the irregularly sampled data to a virtual planar monostatic array. 
However, this efficient multi-planar multistatic (EMPM) algorithm suffers from defocusing due to assumptions in the derivation compared to the BPA when the samples are taken across a large volume.
Additionally, both the BPA and EMPM suffer under errors in position estimation, a phenomenon not thoroughly addressed in previous literature.

To overcome limitations in the BPA and EMPM for irregular arrays under positioning error, we introduce the first convolutional neural network based super resolution (SRCNN) algorithm for near-field freehand SAR imaging. 
The proposed technique involves a two-step alternative which provides competitive results in comparison to BPA, without the time-expensive computation. 
The two-step approach uses the EMPM image reconstruction algorithm \cite{smith2022efficient} followed by a super resolution convolutional neural network (SRCNN) to address SAR image distortion caused by positioning errors, undersampling, and resolution deficiencies of the radar image.
Our proposed algorithm achieves high-resolution, high-efficiency radar images from near-field irregular SAR, such as the case for freehand imaging.

Previous studies on developing CNN-based super resolution networks for near field radar imaging are limited by their datasets of exclusively randomly generated point scatterers \cite{gao2018enhanced,jing2022enhanced}. 
In addition, prior approaches require precise mechanical systems for scanning the radar images and are not suitable for freehand irregular SAR. 
Our approach, enables efficient high-resolution freehand imaging by training on sophisticated, realistic targets and compensating for distortion due to positional errors present in freehand imaging. 
Also, convolutional network based super-resolution methods are bound by their computational complexity due to large number of trainable parameters. 
Toward addressing this drawback, we implement a depthwise separable convolution technique \cite{guo2019depthwise,chollet2017xception} for reducing memory requirements and the training/inference time while maintaining the same image quality.

\begin{figure}[ht]
    \centering
    \includegraphics[scale=.65]{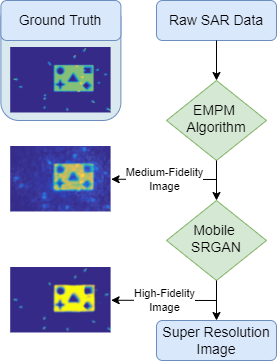}
    \caption{Flowchart of proposed design. After the raw SAR data are processed by the EMPM algorithm \cite{smith2022efficient}, the Mobile-SRGAN is applied to achieve a super resolution image.}
    \label{fig:flowchart}
\end{figure}

The proposed algorithm is summarized if Fig. \ref{fig:flowchart}. 
The raw freehand SAR data are processed by the EMPM algorithm \cite{smith2022efficient} yielding a medium-fidelity image suffering from some image distortion due to positional errors and small array perturbation assumptions in the EMPM.
The example image in Fig. \ref{fig:flowchart} consists of a solid object with some cavities and randomly placed point targets. 
The proposed Mobile-SRGAN processes the distorted image reconstructed using the EMPM and outputs an enhanced, super resolution image.

The main contributions of this paper include:

\begin{itemize}[noitemsep,topsep=1pt]
\item[1)] Generation of near-field SAR training datasets with sophisticated targets 
\item[2)] Development of efficient Super-Resolution Convolutional Neural Network (SRCNN) models on the complex shapes and realistic datasets
\item[3)] Evaluation of the SRCNN models on both real and synthetic images for accuracy and performance metrics
\end{itemize}

The remainder of the paper is formatted as follows. 
In Section \ref{sec:background_dsp}, the requisite radar signal processing topics are briefly discussed. 
Section \ref{sec:mob-srgan} details the design and implementation of the proposed Mobile-SRGAN algorithm.
Simulation and empirical results are discussed in Section \ref{sec:results} followed finally by conclusions.

\section{Preliminaries of Near-Field Freehand SAR}
\label{sec:background_dsp}
Freehand imaging requires efficient data collection, high-fidelity positioning, and efficient image reconstruction for mobile applications.
To achieve efficient sampling, multiple-input multiple-output (MIMO) arrays are commonly employed in the time-division multiplexing (TDM) fashion such that each transceiver pair is activated independently. 
However, the multistatic nature of MIMO arrays complicate the imaging process \cite{yanik2019sparse}. 
The received signal of a MIMO array with transmitter (Tx) and receiver (Rx) antenna locations at $\bm{x}_T$ and $\bm{x}_R$, respectively, can be written as
\begin{equation}
    \label{eq:received_general}
    s(\bm{x}_T,\bm{x}_R,k) = \iiint \frac{o(\bm{x})}{R_T R_R} e^{-jk(R_T + R_R)} d\bm{x},
\end{equation}
where $k$ is the wavenumber of the signal, $o(\cdot)$ is the reflectivity function of the target, $\bm{x}$ are the spatial locations of the target, and $R_T$, $R_R$ are given by
\begin{align}
\label{eq:Rl_of_xT_xR_yT_yR}
        R_T = |\bm{x}_T - \bm{x}|, \quad
        R_R = |\bm{x}_R - \bm{x}|.
\end{align}

Recovering the image of the target requires inverting (\ref{eq:received_general}) to isolate the reflectivity function; however, assuming a freehand scenario, the sample locations $\bm{x}_T$ and $\bm{x}_R$ are captured across multiple planes. 
Hence, the popular RMA \cite{yanik2019sparse} cannot recover the image and though the BPA is suitable for such geometries, it is computational prohibitive. 
To overcome these algorithm deficiencies, the EMPM \cite{smith2022efficient} proposed projecting the multi-planar multistatic samples to a virtual planar monostatic scenario at locations $\bm{x}'$ as
\begin{equation}
    \label{eq:multiplanar_compensation}
    \hat{s}(\bm{x}_\ell',f) \approx s(\bm{x}_T,\bm{x}_R,k) e^{j k\beta_\ell},
\end{equation}
with
\begin{equation}
    \beta_\ell = 2 d_\ell^z + \frac{(d_\ell^x)^2 + (d_\ell^y)^2}{4 Z_0},
\end{equation}
where $\bm{x}_\ell'$ is the location of the $\ell$-th virtual planar monostatic antenna, $Z_0$ is the reference plane, $d_\ell^x$ and $d_\ell^y$ are the distances between the Tx and Rx elements along the $x$- and $y$-directions, respectively, and $d_\ell^z$ is the distance from the Tx/Rx plane to the reference plane. 
By removing the residual phase $\beta_\ell$, the multi-planar multistatic samples are approximately projected to virtual planar monostatic samples and the image can be efficiently recovered by applying the RMA to (\ref{eq:multiplanar_compensation}).
A detailed derivation and explanation the EMPM algorithm can be found in \cite{smith2022efficient}. 

Another crucial element of freehand imaging is position estimation. 
Previous research has employed infrared cameras \cite{alvarez2019freehand} or stereo camera sensors \cite{alvarez2021towards} to achieve precise localization for freehand imaging. 
However, error is observed for either positioning modality that causes aberrations in the recovered image, for both the BPA \cite{alvarez2019freehand,alvarez2021towards} and efficient multi-planar multistatic \cite{smith2022efficient}. 
\begin{figure}[b]
    \centering
    \includegraphics[scale=0.243]{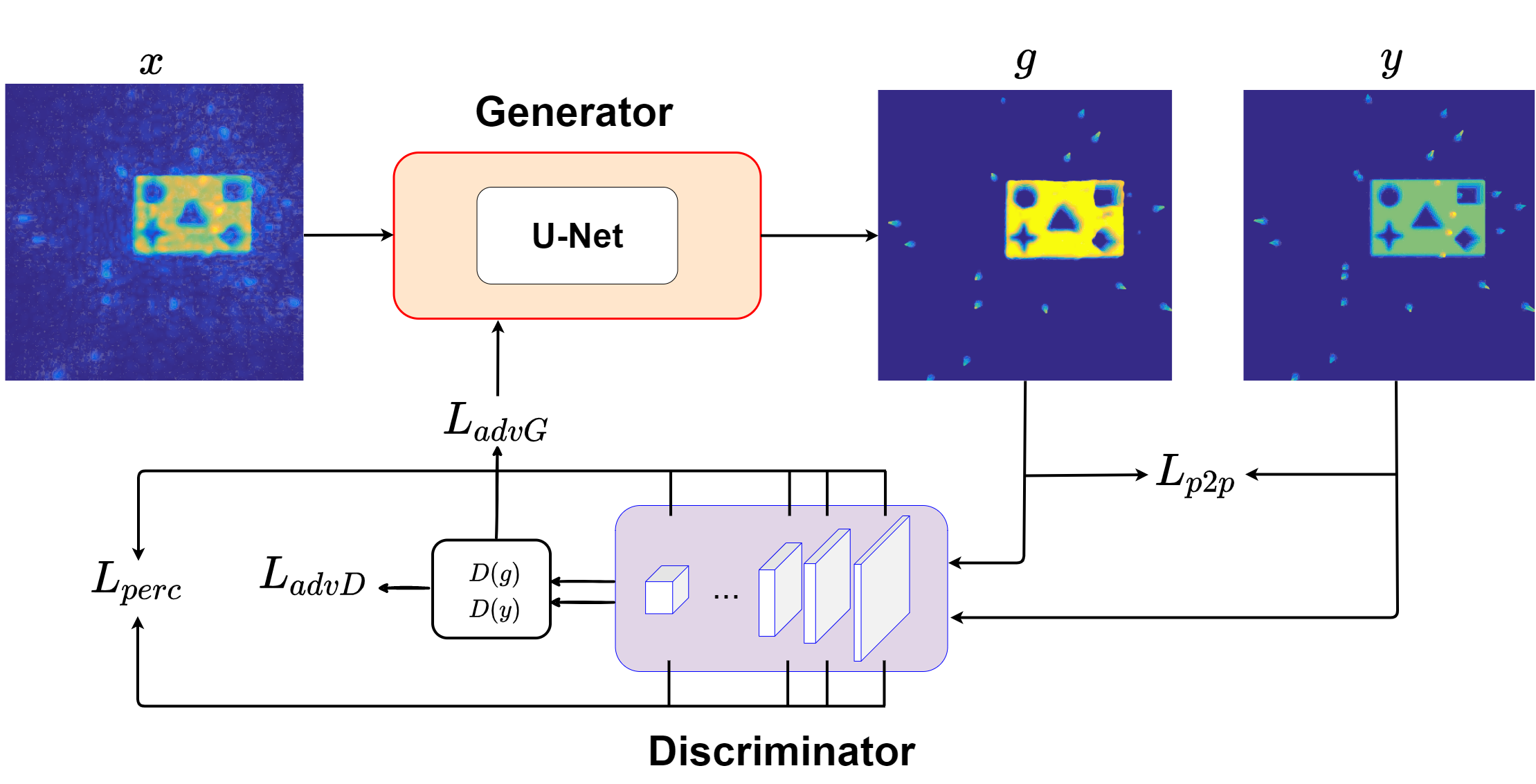}
    \caption{Mobile-SRGAN Overview}
    \label{fig:gan_overview}
\end{figure}
Though positioning errors that cause image blurring and defocusing are unaddressed in in previous work, we consider the case of array position errors in the subsequent analysis.
In this paper, we employ the EMPM to achieve medium-fidelity images and propose the first SAR image super-resolution algorithm for near-field freehand scenarios applicable for mobile applications. 
Our proposed algorithm satisfies the requirements of freehand imaging by employing a MIMO-SAR freehand data collection technique, compensating for position estimation errors, and enabling efficient and mobile-friendly image recovery while achieving higher image quality and resolution than the BPA.


\RestyleAlgo{ruled}
\IncMargin{0em}
\begin{algorithm}[t]
\DontPrintSemicolon
\caption{Training Process for Mobile-SRGAN}
\label{alg:one}
\KwData{$D_{train}, D_{test}$ \hfill\tcp{Train-Test Set}}
\KwData{$x$, $g$ \hfill\tcp{in and out of G}}
\KwData{$y$ \hfill\tcp{HR target images}}
\KwResult{Trained Model}
    \nl\For{$epoch \leftarrow 0 $ \KwTo $ MaxEpochs$}{
        \nl\For{$x \leftarrow 0 $ \KwTo $ batches $ in $ D_{train}$}{
            \nl $g \longleftarrow G(x)$\;
            \nl \tcc{Train Discriminator}
            \nl $L_{advD} \longleftarrow log(D(y))+log(1-D(g))$\;
            \nl $L_{perc} \longleftarrow \sum_{i=0}^{L_D}\lVert F_{D_i}(y) - F_{D_i}(g)\rVert_{1}$\;
            \nl backprop D       \hfill\tcp{Update D}
            \nl \tcc{Train Generator}
            \nl $L_{advG} \longleftarrow log(D(g))$\;
            \nl $L_{p2p} \longleftarrow \lambda_{p2p}\lVert y-g \rVert_{1}$\;
            \nl backprop G       \hfill\tcp{Update G}
        }
        \nl\tcc{Evaluate Generator}
        \nl PSNR$(G(D_{test}),y_{D_{test}})$\;
        \nl RMSE$(G(D_{test}),y_{D_{test}})$\;
    }
    \nl return $G$\;
\end{algorithm}
\DecMargin{0em}


\section{Mobile Super-Resolution GAN}
\label{sec:mob-srgan}

With the reconstructed SAR image from the EMPM, we are tasked with clearing the distortions on these images to generate a super-resolution version that is reflective of the original target image. 
For this purpose, we propose a cGAN-based \cite{goodfellow2014generative, mirza2014conditional} Mobile-SRGAN framework which trains an encoder-decoder model to aggregate features for denoising, thereby generating a high quality radar image. 
This framework consists of two CNNs: the Generator (G) and the Discriminator (D) which work in tandem to optimize the model for achieving super-resolution. 
This particular training framework has been previously used on medical MRI imaging \cite{Armanious2020MedGAN}, super-resolution techniques \cite{Armanious2019An_adversarial_SR} and Image-to-Image translation \cite{Wang2018PAN, isola2018imagetoimage} and we adapt this architecture with depth-wise convolution operations to develop an efficient methodology suitable for mobile applications. 
The Mobile-SRGAN framework is illustrated in Fig. \ref{fig:gan_overview}

The training dataset consists of the synthetic (reconstructed) SAR images and their corresponding ideal images, generated using the methodology in \cite{gao2018enhanced,smith2021An}, provided to the generator-discriminator pair. 
The objective of the generator is to map the distribution of the low-resolution SAR image to the high-resolution ideal image by learning to bridge the gap between those distributions. 
This learning is guided by the discriminator, which is a classifier established to quantify the probabilistic differences between the ``clean'' image ($g$) produced by the generator and the ideal target ($y$). 
The generator attempts to generate enhanced images by transforming the reconstructed SAR image which is then evaluated by the discriminator. 
This information is then fed back to the generator to guide the training toward the distribution of the ideal image. 
The network is trained in an adversarial manner, as shown in Algorithm \ref{alg:one}, until the discriminator is unable to distinguish between the ideal and clean images, implying the generator has adequately learned to generate images akin to the high-resolution images.

The adversarial training for the generator-discriminator architecture is based on the Binary Cross-Entropy Loss ($BCE$) as described in equations \eqref{five} and \eqref{eight}.
The loss terms, $L_{adv}$, $L_{perc}$ and $L_{p2p}$, are combined to train the generator and the discriminator as illustrated in Fig. \ref{fig:gan_overview}.
The role of the discriminator is to differentiate between the clean and the ideal images based on their features, thereby labelling them as clean (0) or ideal (1), using the perceptual loss as calculated in \eqref{six}.
Additionally, the pixel-wise loss term penalizes the discrepancy between the exported image and the clean target \eqref{seven}.

\begin{gather}
    L_{advD} = BCE(D(y), 1) + BCE(D(g), 0), \label{five}\\
    L_{perc}= \sum_{i=0}^{L_D}\lVert F_{D_i}(y) - F_{D_i}(g)\rVert_{1}, \label{six}\\
    L_{p2p}= \lambda_{p2p}\lVert y-g \rVert_{1}, \label{seven}\\
    L_{advG} = BCE(D(g), 1), \label{eight}
\end{gather}

where $F_{D_i}$ is the intermediate features extracted from the $i^{ith}$ convolution layer of the discriminator, $L_D$ is the number of the convolution layers existing in the discriminator, and $\lambda_{p2p}$ is a constant which weights the contribution of the discrepancy. 

\begin{figure*}[ht]
    \centering
    \includegraphics[scale=1, width=0.9\textwidth]{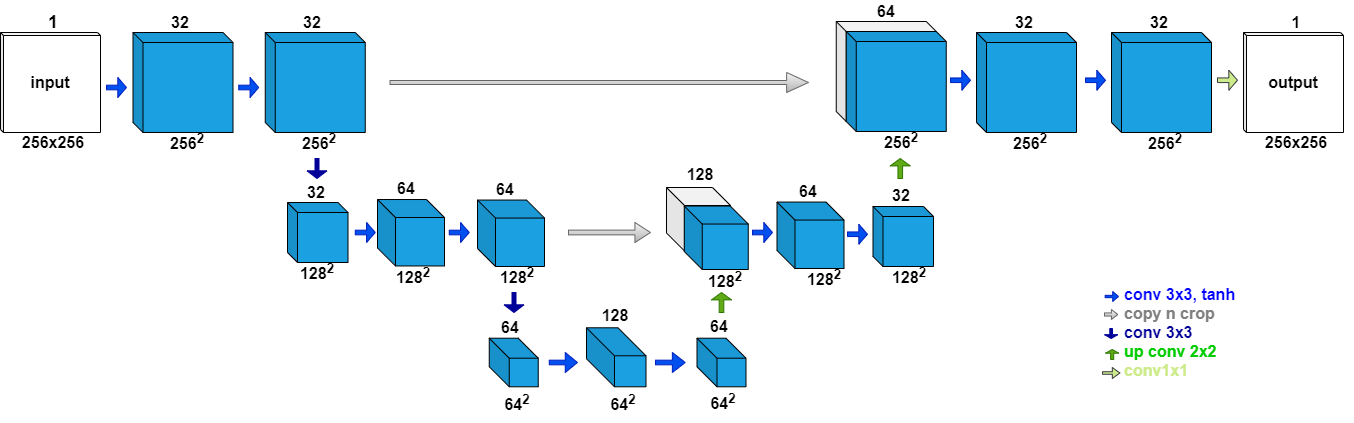}
    \caption{Mobile-SRGAN Generator Architecture} 
    \label{fig:mobu-net}
\end{figure*}

We implemented the generator using an encoder-decoder variant \cite{ronneberger2015unet}, as illustrated in Fig. \ref{fig:mobu-net}, applying a spatial reduction step after every convolutional bottleneck on the decoder and a resolution recovering step in the encoder. 
On each bottleneck's output, a corresponding residual connection forwards the needed features, avoiding loss of information. 
The novelty of this encoder-decoder architecture is that it factorizes the standard $3 \times 3$ convolution into a $3 \times 3$ depthwise convolution (DWC) and a $1 \times 1$ pointwise convolution (PWC) to drastically improve efficiency \cite{guo2019depthwise}. 

DWC and PWC play different roles in generating new features: the former captures spatial correlation information while the latter measures the channel-wise correlations.
This results in a substantial reduction to the CNN model size, with up to 75\% reduction by applying a DWC on the first layer consisting of 32 feature channels. 
Hence, the proposed Mobile-SRGAN architecture, with 79,233 model parameters, is suitable for many mobile and smartphone applications \cite{howard2017mobilenets}.
By incorporating the DWC, a resulting parameterization of the spatial reduction step occurs. 
This influences the performance, letting the training process utilize the spatial reduction as desired.
The discriminator is based on the patch discriminator architecture introduced by \cite{isola2018imagetoimage}, which divides the input into $16 \times 16$ patches and classifies each patch. 
The classification score is averaged out among all patches \cite{Armanious2020MedGAN}. 

The efficient mobile SAR super resolution framework was trained for 50 epochs using the ADAM optimizer on a single TESLA P100 GPU with 16 GB using 4096 samples for the training process and 1024 for the evaluation. 
The training dataset consists of SAR images containing randomly placed point scatterers, solid objects, and hollow objects, an improvement over previous work which only includes point scatterers \cite{gao2018enhanced,jing2022enhanced,smith2021An}.
The training process was approximately 8 and a half hours with an inference time during validation process of 14 ms per sample.

\begin{figure}[h]
  \centering
  \resizebox{\columnwidth}{!}{%
    \begin{tabular}{ccc}
    \includegraphics[width=0.33\linewidth]{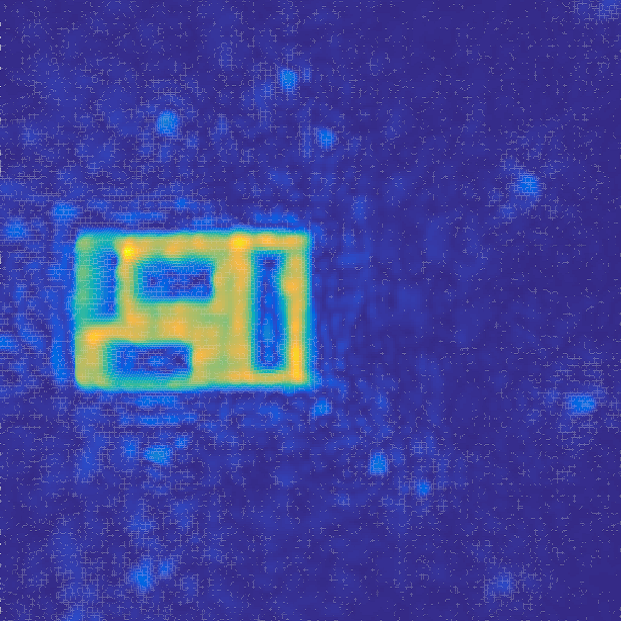}    
    &\includegraphics[width=0.33\linewidth]{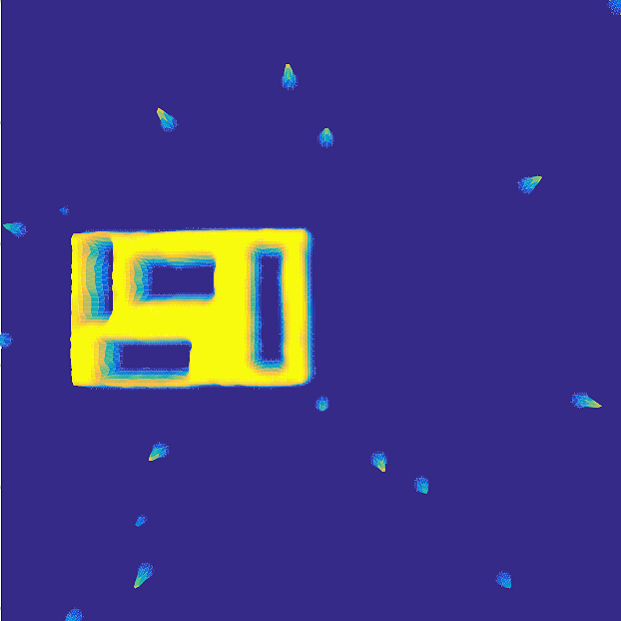}
    &\includegraphics[width=0.33\linewidth]{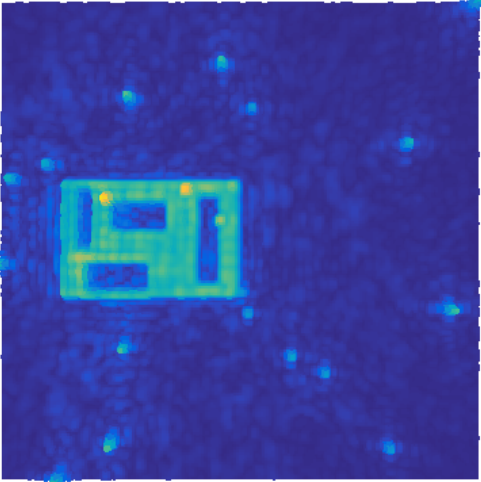}\\
    \includegraphics[width=0.33\linewidth]{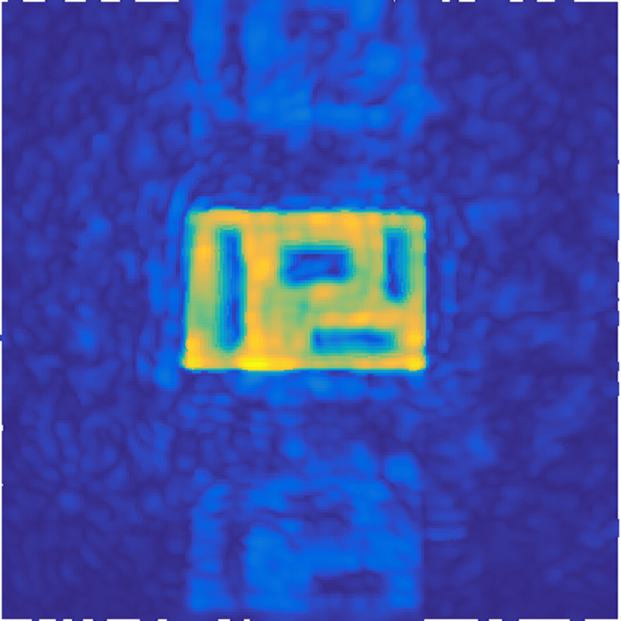}
    &\includegraphics[width=0.33\linewidth]{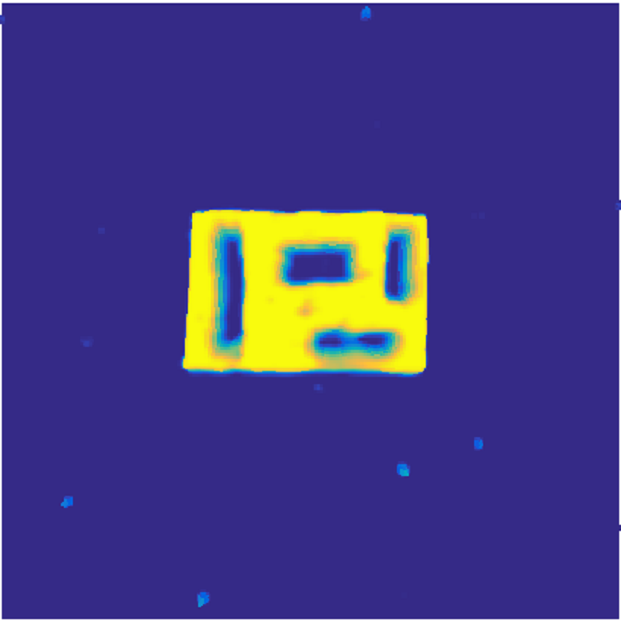}
    &\includegraphics[width=0.33\linewidth]{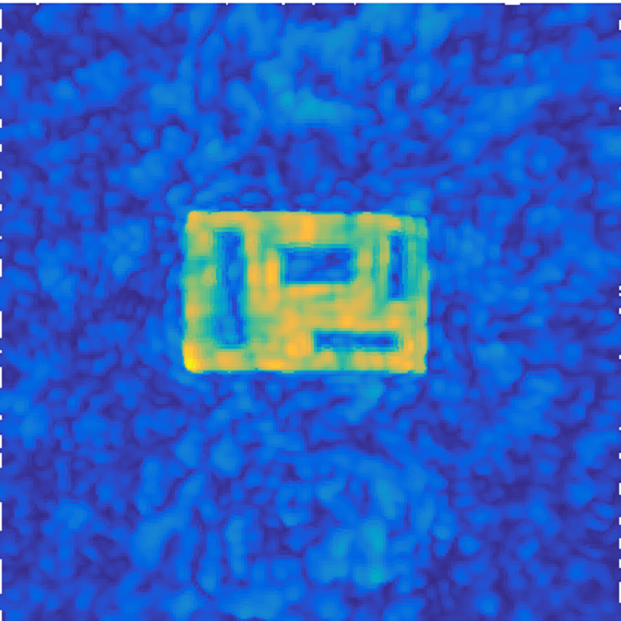}
    \\
    (a) EMPM & (b) Mobile-SRGAN & (c) BPA
    \end{tabular}
  }
\caption{Qualitative Results: Comparison of clean images (a) EMPM Algorithm, (b) Mobile-SRGAN (CNN-based Super Resolution Algorithm), and (c) BPA Algorithm. Results from synthetic images are shown on the first row, while results from real images are shown in the second row.}
  \label{fig:qualitative_res}
\end{figure}

\section{Results}
\label{sec:results}
We evaluate the adversarial framework with a test set consisting of 1027 ``never-seen-before'' images (1024 synthetic images and 3 real images) excluded during the model training.
The qualitative results are shown in Fig. \ref{fig:qualitative_res}. 
The network is capable of producing rigid high-resolution images when processing new low-resolution data. 
Through learning the joint distribution mapping between low-high resolution images, the generator is able to create consistent and well-structured objects by mostly removing the existing distortion around the image while enhancing the appropriate local structure and identifying the fingerprint of the objects. 
The quantitative performance of the proposed framework can be measured by computing the peak signal-to-noise ratio (PSNR) and the root mean square error (RMSE) between the generated images and ideal images of the synthetic testing dataset.
As shown in Table \ref{table:Table1}, the proposed method outperforms the existing methods in terms of both computational efficiency and image quality. 
Particularly, our algorithm achieves higher PSNR and lower RMSE than the gold-standard BPA with only a slight increase in computation time for a single image compared to the EMPM and RMA. 


\begin{table}[h!]
\centering
\begin{tabular}{ |c|c|c|c|c| } 
\toprule
\textbf{Metrics} & \textbf{Mobile-SRGAN} & \textbf{BPA} & \textbf{EMPM} & \textbf{RMA} \\ [0.5ex] 
\midrule
PSNR (dB) & $\mathbf{34.926}$ & $26.33$ & $20.20$ & $10.158$\\ 
RMSE & $\mathbf{0.019}$ & $0.044$ & $0.105$ & $0.276$\\
Time (s) & $1.117$ & $1324.8$ & $\mathbf{1.103}$ & $\mathbf{1.103}$\\
\bottomrule
\end{tabular}
\caption{Quantitative performance of the Mobile-SRGAN compared to the BPA, EMPM, and RMA.} 

\label{table:Table1}
\end{table}

\section{Conclusion}
\label{sec:conclusion}
In this paper, we introduce the first CNN-based super resolution algorithm for mobile freehand SAR imaging in the near-field. 
Rather than training on randomly placed point targets, we improve upon previous work by incorporating solid, intricate objects in the simulation that are more representative of real-world scenarios. 
The proposed CNN algorithm is applied to images recovered by the EMPM algorithm \cite{smith2022efficient} yielding high-resolution low-noise SAR images and outperforming previous techniques. 
The Mobile-SRGAN is the first SAR super resolution algorithm developed for freehand radar imaging, a more difficult task than traditional SAR super resolution, and efficiently recovers high-resolution images with low computational cost, deeming it fit for computationally constrained applications. 

\bibliographystyle{IEEEbib}
\bibliography{mega_bib}

\end{document}